\documentclass[pdflatex,sn-mathphys-num]{sn-jnl}

\usepackage{graphicx}%
\usepackage{multirow}%
\usepackage{amsmath,amssymb,amsfonts}%
\usepackage{amsthm}%
\usepackage{mathrsfs}%
\usepackage[title]{appendix}%
\usepackage{xcolor}%
\usepackage{textcomp}%
\usepackage{manyfoot}%
\usepackage{booktabs}%
\usepackage{algorithm}%
\usepackage{algorithmicx}%
\usepackage{algpseudocode}%
\usepackage{listings}%
\usepackage{longtable}%
\usepackage{booktabs}%
\usepackage{multirow}
\usepackage{caption}
\usepackage{booktabs}
\usepackage{subcaption}   
\theoremstyle{thmstyleone}%

\theoremstyle{thmstyletwo}%

\theoremstyle{thmstylethree}%

\begin{document}

\title[FM Validation]{\center Validation of Whole-Slide Foundation Models \\for Image Retrieval in TCGA Data}

\author[1,2]{\fnm{Tianhao} \sur{Lei}}
\equalcont{These authors contributed equally to this work.}

\author[1,3]{\fnm{Parsa} \sur{Esmaeilkhani}}
\equalcont{These authors contributed equally to this work.}

\author[1]{\fnm{Saghir} \sur{Alfasly, PhD}}

\author[1]{\fnm{Wataru} \sur{Uegami, MD,PhD}}

\author[4]{\fnm{Judy C.} \sur{Boughey, MD}}
\author[45]{\fnm{Matthew P.} \sur{Goetz, MD}}
\author[5]{\fnm{Krishna R.} \sur{Kalari, PhD}}
\author*[1]{\fnm{H.R.} \sur{Tizhoosh, PhD}}\email{tizhoosh.hamid@mayo.edu}

\affil[1]{KIMIA Lab, Artificial Intelligence and Informatics, Mayo Clinic, Rochester, MN, USA}

\affil[2]{Department of Neurology, Northwestern University Feinberg School of Medicine, Chicago, IL, USA}

\affil[3]{Department of Computer Science, Temple University, Philadelphia, PA, USA}

\affil[4]{Department of Breast and Melanoma Surgical Oncology, Comprehensive Cancer Center, Mayo Clinic, Rochester, MN, USA}
\affil[5]{Department of Oncology, Comprehensive Cancer Center, Mayo Clinic, Rochester, MN, USA}
\affil[6]{Department of Quantitative Health Sciences, Mayo Clinic, Rochester, MN, USA}

\abstract{Foundation models are reshaping computational histopathology, yet their value for whole-slide image retrieval remains uncertain, particularly relative to strong patch-based and supervised aggregation baselines. Here, we benchmarked ten retrieval pipelines on 9,387 diagnostic slides spanning 17 organs and 60 organ-specific diagnoses from The Cancer Genome Atlas (TCGA) under patient-level leave-one-patient-out evaluation. Pipelines include four pre-trained slide foundation models, a supervised attention-based multiple instance learning (ABMIL) aggregator built on patch embeddings, and a patch-level retrieval paradigm evaluated across five sampling densities. Retrieval performance varies far more across organs and diagnostic categories than across model architectures. Although the pre-trained slide foundation model TITAN achieved the strongest overall performance, its advantage was modest; ABMIL and patch-based approaches reached comparable accuracy under Top-1 and Top-3 retrieval, with no single model consistently dominant across all organs or diagnoses. Morphologically distinctive entities approached ceiling retrieval across methods, while rare, heterogeneous, and closely related subtypes remained consistently challenging. Critically, misclassifications aligned closely with organs exhibiting known inter-observer variability among expert pathologists, suggesting an intrinsic ceiling for morphology-only retrieval. 
Importantly, retrieval quality appears to be driven primarily by patch-level feature representations, with slide-level aggregation contributing limited additional benefit, suggesting that aggregation may not be necessary for strong performance in many settings. Taken together, these findings argue against a universally optimal retrieval architecture and instead motivate organ-resolved benchmarking, diagnosis-aware or ensemble deployment strategies, continued investment in stronger feature representations, and multimodal retrieval frameworks. More fundamentally, even the strongest model achieves only $\approx 68\% \pm 21\%$ retrieval accuracy on TCGA, and for several diagnostic subtypes all evaluated models fail entirely ($0\%$ retrieval accuracy), underscoring that despite architectural advances, substantial progress is still required before such systems reach reliable clinical utility and revealing fundamental limitations in current morphology-based representations.}

\maketitle

\section{Introduction}\label{sec1}
The introduction of computer vision into histopathology has transformed tissue sample analysis, enabling computational approaches that complement and, in selected settings, match or exceed human-level performance in tasks ranging from cancer detection and subtyping to biomarker prediction and prognostication~\cite{coudray_classification_2018,lu_data-efficient_2021,chen_towards_2024,campanella_clinical_2025}. Central to this transformation are \emph{foundation models} (FMs): large neural networks pretrained on large-scale histopathology datasets using self-supervised learning (SSL) and/or vision--language alignment, which yield general-purpose feature representations transferable to diverse downstream clinical tasks with minimal task-specific adaptation~\cite{chen_towards_2024,vorontsov_foundation_2024,xu_multimodal_2025,ding2025multimodal,lu_visual-language_2024}. Over the past two years, the field has witnessed rapid proliferation of histopathology FMs~\cite{chen_towards_2024,vorontsov_foundation_2024,lu_visual-language_2024,wang_transformer-based_2022,xu_whole-slide_2024,shaikovski_prism_2024,ding2025multimodal,xiong_survey_2025,leonardis_multistain_2025}. Some models operate in a two-tier paradigm. First, patch-level encoders---often vision transformers trained with SSL or vision-language objectives on millions of tiny histology tiles---produce embeddings for individual image patches. Second, slide-level models aggregate the variable-length set of patch embeddings from a whole-slide image (WSI) into a fixed-dimensional slide representation using learned aggregation mechanisms such as supervised attention-based multiple instance learning (ABMIL), cross-attention distillation, or transformer-based spatial modeling~\cite{ilse_attention-based_nodate,lu_data-efficient_2021,shaikovski_prism_2024,ding2025multimodal,xu_whole-slide_2024}. These slide-level models vary widely in architecture, pretraining strategy, and dataset composition ranging from tens of thousands to hundreds of thousands of WSIs, yet share a common goal: to provide general-purpose backbones for computational pathology. A recent example is TITAN~\cite{ding2025multimodal}, a bimodal whole-slide foundation model pretrained on 335{,}645 WSIs using visual self-supervision and vision--language alignment with pathology reports.

The rapid proliferation of pathology FMs has motivated several independent benchmarking efforts. Neidlinger et al. compared 19 patch encoders across 31 weakly supervised tasks spanning morphology, biomarker and prognostic prediction \cite{neidlinger_benchmarking_2025}. Their analysis identified strong performers including CONCH v1.5, and demonstrated that patch-level models trained on distinct cohorts encode complementary signals amenable to fusion~\cite{neidlinger_benchmarking_2025}. Campanella et al. assessed public self-supervised encoders across three medical centers, observing consistent improvements over ImageNet baselines but largely modest task-dependent differences among leading models \cite{campanella_clinical_2025}. Chen et al. benchmarked embedding aggregation strategies, finding that no single slide-level aggregator dominates across tasks \cite{chen_towards_2024}. Broader surveys further underscore the diversity of model architectures and evaluation protocols, as well as the lack of standardization across studies \cite{xiong_survey_2025}. Critical perspectives have additionally argued that current FMs remain misaligned with practical pathology needs, citing low retrieval accuracy, fragile cross-institutional generalization and sensitivity to routine imaging variation as persistent limitations \cite{tizhoosh_beyond_2025}. Collectively, these studies establish important baselines for supervised prediction, yet they share a common dependence on task-specific training—whether through linear probing, attention-based aggregation or full fine-tuning—over frozen or adaptable foundation model embeddings \cite{chen_towards_2024,ding2025multimodal,neidlinger_benchmarking_2025,campanella_clinical_2025,xu_whole-slide_2024}. The intrinsic quality of learned representations in a strictly zero-shot, label-free setting remains less systematically characterized.

This gap is consequential because WSI retrieval can be formulated as a purely zero-shot problem: a query slide is matched against a large archive to retrieve morphologically or diagnostically similar cases using pretrained embeddings alone, without task-specific optimization or parameter updates~\cite{kalra_yottixel_2020,kalra_pan-cancer_2020,tizhoosh_beyond_2025,lahr_analysis_2025}. Such a paradigm provides a stringent test of representation quality, probing the intrinsic discriminative structure of the embedding space rather than the capacity of a downstream classifier to compensate for representational deficiencies. Alfasly et al.\cite{alfasly_validation_2025} made an important step in this direction, evaluating three FMs---UNI, Virchow and GigaPath---alongside a DenseNet baseline within the Yottixel retrieval framework on The Cancer Genome Atlas Program (TCGA). Reporting macro-averaged F1 scores across 23 organs and 117 cancer subtypes, they found that retrieval accuracy remains well below clinical thresholds, with pronounced organ-level variability. However, the scope of that study was understandably limited: only a small subset of available FMs was assessed while new models emerged (CONCH, TITAN, among others), and slide-level aggregation was represented by a single architecture (GigaPath WSI). A comprehensive zero-shot retrieval benchmark encompassing a substantially wider range of both patch-level and slide-level FMs, evaluated under a unified protocol with macro-averaged F1 as the primary metric, has not yet been reported. Beyond this central gap, several further limitations motivate the present work.

First, existing evaluations typically emphasize aggregate performance across tasks or organs, providing limited insight into organ- and diagnosis-specific behavior. Given the heterogeneity of tissue morphology across anatomical sites, no single model is expected to dominate universally~\cite{alfasly_validation_2025,tizhoosh_beyond_2025}; however, the pattern and extent of complementary strengths among FMs remain poorly characterized at organ and diagnosis resolution. Understanding which model excels in which diagnostic context--and where failure modes concentrate--is essential for informing clinical deployment strategies, including model selection and principled ensembling.

Second, the marginal contribution of learned slide-level aggregation to retrieval performance remains an open question. Slide-level FMs compress thousands of patch embeddings into a single slide vector through pretrained aggregation architectures~\cite{ding2025multimodal,shaikovski_prism_2024,leonardis_multistain_2025,xu_whole-slide_2024}. If we do not use aggregation-based methods, patch embeddings produced by FMs must be evaluated within a reliable and adaptable search framework. Such a search engine should satisfy several key requirements: (1) it must allow seamless integration of different FMs without necessitating significant architectural changes or extensive ablation studies; (2) it must employ an unsupervised patch-selection strategy, thereby eliminating the need for retraining when extending the search to additional organs or tissue types; (3) it should support WSI retrieval by leveraging all selected patches without requiring an explicit aggregation step; and (4) it must demonstrate storage efficiency, fast search performance, and high retrieval accuracy compared with alternative solutions. An alternative, \emph{aggregation-free} paradigm is exemplified by Yottixel~\cite{kalra_yottixel_2020}, which selects a diversified subset of representative patches per WSI via chromatic and spatial clustering and computes slide-to-slide similarity directly from patch-level distances using a median-of-minimum patch comparison  scheme~\cite{kalra_pan-cancer_2020,alfasly_validation_2025}. This preserves local morphological correspondence and avoids potential information loss from collapsing heterogeneous tissue into a single vector. Yottixel’s flexible topology enables seamless integration of diverse deep learning models alongside unsupervised patch-selection strategies. This flexibility made it particularly well suited for our retrieval experiments, as it preserved architectural integrity while accommodating new models like CONCH and patch-selection methods without modification. Although Alfasly et al. compared GigaPath WSI against patch-level baselines and found limited benefit from aggregation~\cite{alfasly_validation_2025}, whether this finding generalizes across a broader set of slide-level architectures---and whether the additional complexity and pretraining cost of such aggregators translate into meaningfully improved retrieval accuracy when paired with stronger patch encoders--remains unresolved.

We present, to our knowledge, the first large-scale evaluation of end-to-end WSI retrieval pipelines using slide-level FMs across TCGA (pervious studies have used patch comparisons only \cite{kalra_pan-cancer_2020}). We benchmarked ten retrieval paradigms on 9{,}387 diagnostic slides spanning 17 organ types and 60 organ-specific diagnostic categories, using macro-averaged F1 score under leave-one-patient-out  evaluation as the primary metric. These paradigms comprise five slide-level models--four pre-trained whole-slide FMs (TITAN, PRISM, Madeleine and GigaPath) and one ABMIL trained on CONCH v1.5 patch embeddings--together with a patch-level retrieval framework (Yottixel with CONCH v1.5 embeddings at four sampling rates: 5\%, 10\%, 20\%, and 50\%, as well as using all the patch embeddings). Our analysis characterizes performance at organ and diagnosis resolution, quantifies structured misclassification patterns, assesses the relationship between diagnostic prevalence and retrieval accuracy, and examines the relative contribution of learned slide-level aggregation versus patch-level retrieval. Critically, by including ABMIL alongside pre-trained slide-level FMs and the aggregation-free Yottixel framework--all using the same CONCH v1.5 patch encoder--we directly assess the marginal benefit of supervised versus unsupervised aggregation and isolate the contribution of learned slide-level representations from patch encoding quality. By evaluating all models in a unified retrieval setting, we provide a clinically grounded assessment of how current FMs and supervised aggregators perform when deployed for diagnostic case matching. This work addresses a critical gap in the computational pathology benchmarking landscape.

\section{Results}\label{sec2}

\subsection{Organ-level retrieval performance is heterogeneous and model-dependent}

\begin{figure}[htbp]
    \centering
    \includegraphics[width=\linewidth]{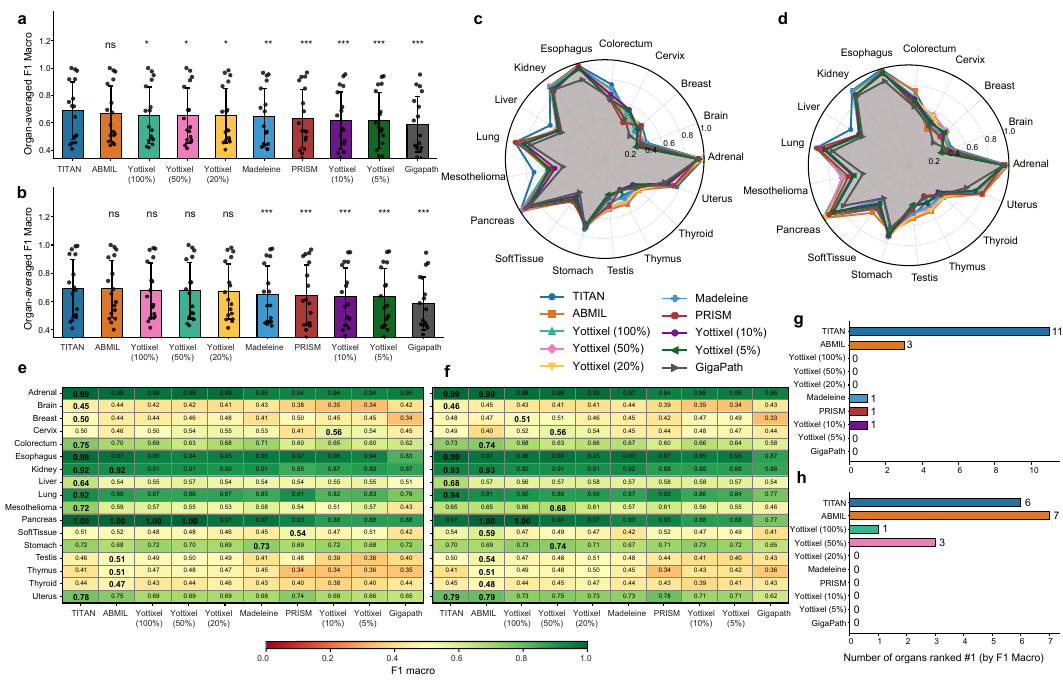}
    \caption{\textbf{Comparative evaluation of whole-slide image retrieval models across 17 organs.} \textbf{a,b}, Mean organ-averaged F1-macro across 17 organs for Top-1 (\textbf{a}) and Top-3 (\textbf{b}) retrieval. Bars denote mean $\pm$ s.d.; dots indicate individual organ scores. Significance brackets indicate two-sided paired comparisons against TITAN (\textit{ns}, not significant; *$p<0.05$; **$p<0.01$; ***$p<0.001$; $n=17$ organs). \textbf{c,d}, Radar plots showing organ-level F1-macro for all ten retrieval configurations under Top-1 (\textbf{c}) and Top-3 (\textbf{d}) retrieval. \textbf{e,f}, Heatmaps of organ-specific F1-macro under Top-1 (\textbf{e}) and Top-3 (\textbf{f}) retrieval; bold values indicate the highest score within each organ. \textbf{g,h}, Number of organs for which each model achieved the highest F1-macro under Top-1 (\textbf{g}) and Top-3 (\textbf{h}) retrieval. [Note that Yottixel+CONCH has been shortened to Yottixel in the figure]}
    \label{fig:overall_performance}
\end{figure}

TITAN achieved the highest organ-averaged F1-macro under Top-1 retrieval ($0.689 \pm 0.209$; $n = 17$ organs), outperforming eight of nine comparators (two-sided paired $t$-test with Holm--Bonferroni correction for multiple comparisons, all adjusted $P \leq 0.033$; Table~\ref{tab:organ_level_performance}, Fig.~\ref{fig:overall_performance}\textbf{a}).

These aggregate rankings, however, mask substantial organ-level heterogeneity (Fig.~\ref{fig:overall_performance}\textbf{c--f}). Organs with morphologically distinctive diagnoses---including pancreas (Top-1 F1 = 1.00 for TITAN and ABMIL), adrenal gland (0.99), esophagus (0.98), and kidney (0.92)---approached ceiling performance across all ten configurations, reflecting the relative ease of separating well-differentiated tissue architectures in embedding space. In contrast, organs characterised by morphological overlap among diagnostic subtypes---thymus (F1 $\approx$ 0.34--0.51), thyroid (0.38--0.47), testis (0.34--0.51), and brain (0.34--0.46)---exhibited broadly low performance, though ABMIL achieved the highest scores in all three of these morphologically challenging organs under both Top-1 and Top-3 retrieval. This pattern indicates that retrieval accuracy is primarily governed by the morphological separability of the diagnostic categories within each organ, with supervised aggregation providing a modest advantage for the most diagnostically complex sites \cite{litjens2017survey,campanella2019clinical,ehteshami2017diagnostic,tizhoosh2021searching}.

Critically, \underline{no single model dominated across all organs}. In the Top-1 setting, TITAN achieved the highest F1-macro in 11 of 17 organs, but the remaining wins were distributed across four other models (Fig.~\ref{fig:overall_performance}\textbf{e,g}): ABMIL ranked first in 3 organs (testis, thymus and thyroid---sites where diagnostic subtype complexity likely benefits from supervised over unsupervised aggregation), while Madeleine, PRISM and Yottixel (10\%) each led in one organ. Notably, patch-based Yottixel variants did not rank first in any organ under Top-1 retrieval. Under Top-3 retrieval, the ranking landscape shifted substantially: ABMIL led 7 of 17 organs (the highest organ-level win count of any model), TITAN ranked first in 6 organs, and Yottixel (50\%) in 3 organs, with Yottixel (100\%)\footnote{Although we used Yottixel with configurations of 50\% and 100\% patch retention, this departs from its original design principle. Yottixel’s mosaic strategy is specifically intended to operate on a small, representative subset of tissue patches, enabling efficient exploration of the clustering space while preserving morphological diversity \cite{kalra_yottixel_2020,kalra_pan-cancer_2020}. The use of full (100\%) patch retention therefore contradicts this intended efficiency-driven design. However, we included both settings for experimental completeness and to facilitate direct comparison with whole-slide image (WSI)–level models, which typically process all available patches during training and inference. In this context, the 100\% retention configuration should be interpreted not as a faithful implementation of the Yottixel paradigm, but rather as a controlled baseline for assessing the impact of patch subsampling on retrieval performance.} adding one further organ win (Fig.~\ref{fig:overall_performance}\textbf{f,h}). The pronounced gain of ABMIL from 3 to 7 organ wins under Top-3 retrieval suggests that within-distribution retrieval might reduce the confusion when multiple nearest neighbours are considered jointly.

Please note that, throughout the results, references to Yottixel correspond to the combined framework Yottixel + CONCH v1.5. This convention is adopted for simplicity and consistency. For brevity, figures and diagrams may also use the shorthand “Yottixel” to denote this combined approach.

\begin{table}[htbp]
\centering
\caption{Organ-level performance comparison of ten retrieval configurations under Top-1 and Top-3 retrieval. Values report mean F1-macro $\pm$ s.d. across 17 organs, with 95\% confidence intervals (CI). TITAN is treated as the baseline model.}
\label{tab:organ_level_performance}
\begin{tabular}{lcccc}
\toprule
\multirow{2}{*}{Model}
& \multicolumn{2}{c}{Top-1 Retrieval}
& \multicolumn{2}{c}{Top-3 Retrieval} \\
\cmidrule(lr){2-3} \cmidrule(lr){4-5}
& Mean $\pm$ s.d. & 95\% CI
& Mean $\pm$ s.d. & 95\% CI \\
\midrule
TITAN
& 0.689 $\pm$ 0.21 & [0.581, 0.796]
& 0.689 $\pm$ 0.21 & [0.583, 0.795] \\

ABMIL
& 0.667 $\pm$ 0.20 & [0.563, 0.772]
& 0.687 $\pm$ 0.20 & [0.582, 0.791] \\

Yottixel+CONCH (100\%)
& 0.656 $\pm$ 0.20 & [0.551, 0.761]
& 0.674 $\pm$ 0.20 & [0.570, 0.778] \\

Yottixel+CONCH (50\%)
& 0.655 $\pm$ 0.20 & [0.552, 0.758]
& 0.674 $\pm$ 0.20 & [0.573, 0.776] \\

Yottixel+CONCH (20\%)
& 0.654 $\pm$ 0.20 & [0.552, 0.756]
& 0.668 $\pm$ 0.20 & [0.567, 0.768] \\

Madeleine
& 0.645 $\pm$ 0.21 & [0.539, 0.752]
& 0.647 $\pm$ 0.20 & [0.542, 0.752] \\

PRISM
& 0.632 $\pm$ 0.21 & [0.524, 0.741]
& 0.643 $\pm$ 0.22 & [0.532, 0.753] \\

Yottixel+CONCH (10\%)
& 0.619 $\pm$ 0.21 & [0.512, 0.726]
& 0.634 $\pm$ 0.20 & [0.529, 0.738] \\

Yottixel+CONCH (5\%)
& 0.618 $\pm$ 0.20 & [0.513, 0.723]
& 0.630 $\pm$ 0.20 & [0.526, 0.734] \\

GigaPath
& 0.590 $\pm$ 0.20 & [0.486, 0.693]
& 0.583 $\pm$ 0.19 & [0.485, 0.681] \\
\bottomrule
\end{tabular}
\end{table}

\subsection{Heterogeneous diagnosis-level retrieval reveals complementary model strengths}

\begin{figure}[htbp]
    \centering
    \includegraphics[width=\linewidth]{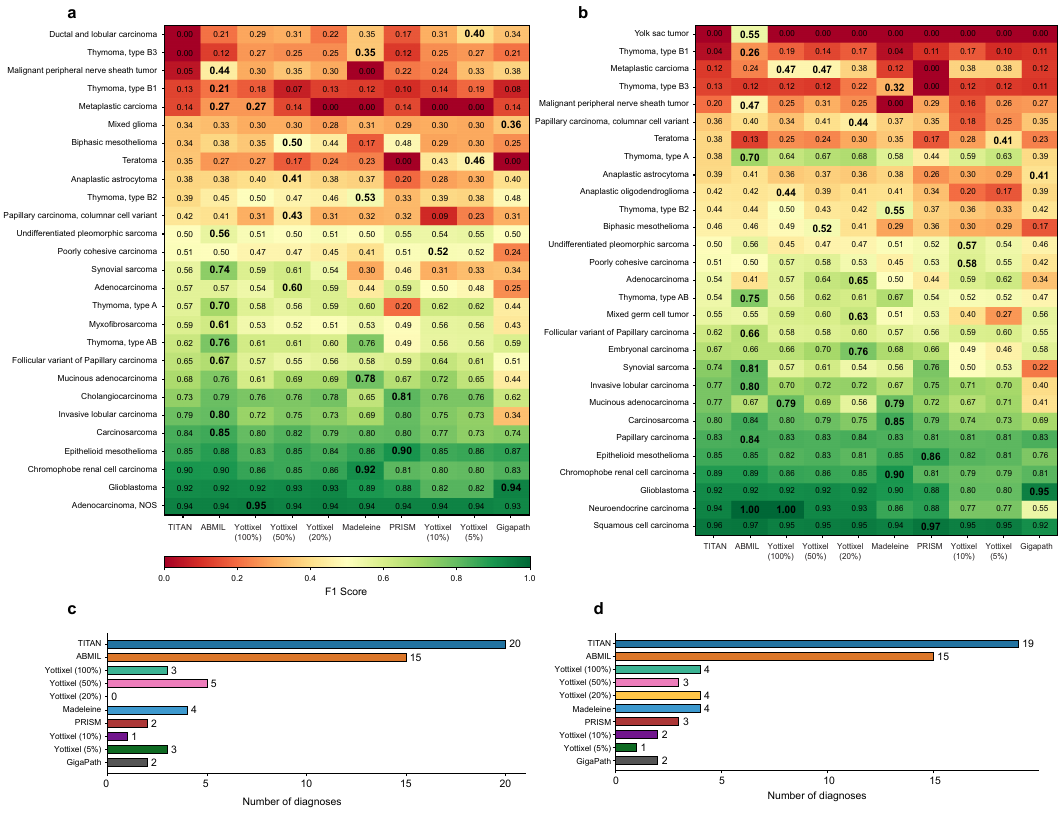}
    \caption{\textbf{Diagnosis-level variation in retrieval performance across ten retrieval configurations.} \textbf{a,b,} Heat maps showing F1 scores under Top-1 retrieval (\textbf{a}) and Top-3 retrieval (\textbf{b}) across TITAN, ABMIL, patch-based Yottixel+CONCH variants (5\%, 10\%, 20\%, 50\%, 100\%), Madeleine, PRISM and GigaPath. Only diagnoses for which TITAN did not achieve the highest F1 score were included (27 diagnoses under Top-1; 29 diagnoses under Top-3). Rows indicate diagnoses and columns indicate models; bold values indicate the highest score within each diagnosis; higher F1 scores are shown in green and lower scores in red. \textbf{c,d,} Number of diagnoses for which each model ranked first by F1 score under Top-1 (\textbf{c}) and Top-3 (\textbf{d}) retrieval (ties at scores of 0 or 1 excluded). [Note that Yottixel+CONCH has been shortened to Yottixel in the figure]}
    \label{fig:diagnosis_level_performance}
\end{figure}

Diagnosis-level analysis reveals substantial heterogeneity in model performance across all ten retrieval configurations but relative model ordering is largely preserved between Top-1 and Top-3 retrieval.(Fig.~\ref{fig:diagnosis_level_performance}\textbf{a,b}; Extended Data Fig.~\ref{fig:f1_diagnosis_breakdown}). Across the diagnostic categories, F1-scores span a broader range than at the organ level, from near-ceiling performance for morphologically distinctive entities (for example, glioblastoma, hepatocellular carcinoma and chromophobe renal cell carcinoma) to uniformly low scores for rare or morphologically overlapping diagnoses (for example, adenosquamous carcinoma and certain thymoma subtypes).  Expanding retrieval from Top-1 to Top-3 modestly increases F1-scores for many intermediate-difficulty diagnoses, but does not substantially alter the overall distribution of strong and weak categories (Extended Data Fig.~\ref{fig:f1_diagnosis_breakdown}). Diagnoses that are well classified under Top-1 generally also remain high-performing under Top-3, whereas categories with persistent confusion remain low across models, indicating that performance limits are driven primarily by intrinsic morphological ambiguity rather than strict nearest-neighbor ranking.

To examine where TITAN does not lead, we visualized F1-scores for diagnoses in which other models achieved the highest F1-score (Fig.~\ref{fig:diagnosis_level_performance}\textbf{a,b}). Three consistent patterns emerge. First, rare composite or histologically heterogeneous entities show low F1-scores across all models, with minimal separation between methods, reflecting shared difficulty rather than model-specific failure. Second, certain closely related subtypes---particularly within thymoma and mixed epithelial malignancies---display model-dependent ranking shifts between Top-1 and Top-3 retrieval, suggesting sensitivity to subtle architectural cues. Third, among high-performing diagnoses (F1 $>$ 0.8), leadership frequently varies by small absolute margins, indicating that multiple models capture largely overlapping but not identical morphological representations. These patterns are mirrored in the cases where TITAN does lead: margin size tracks closely with diagnostic difficulty, narrowing where morphological signals are strong and widening where they are not.

The magnitude of TITAN's advantage varied by diagnosis. For highly recognisable entities, TITAN's lead over the next-best model was narrow but consistent, as seen for \emph{paraganglioma} (\(0.99\) versus \(0.98\)) and \emph{clear cell renal cell carcinoma} (\(0.96\) versus \(0.95\)), suggesting that these diagnoses are retrieved reliably by most models once strong morphological signals are present. By contrast, intermediate-performance diagnoses showed wider inter-model dispersion. For example, in Top-1 retrieval, \emph{serous carcinoma} ranged from \(F1=0.71\) for TITAN to \(0.26\) for the weakest model, and mixed germ cell tumour ranged from \(0.66\) to \(0.46\), indicating that model differences become more consequential in diagnostically less distinct categories. A notable case was combined hepatocellular-cholangiocarcinoma, which appeared only in the Top-3 TITAN-best set: TITAN achieved \(F1=0.29\), whereas all competing models remained at \(0.00\). This result suggests that allowing multiple retrieval candidates can reveal differences in ranking quality that are not apparent under Top-1 evaluation alone.

To complement the diagnosis-level failure analysis, we examined the diagnoses for which TITAN uniquely achieved the highest F1-macro score across all evaluated models (Extended Data Fig.~\ref{fig:titan_best_diagnoses}). TITAN was uniquely best for 12 diagnoses under Top-1 retrieval and 11 diagnoses under Top-3 retrieval, spanning multiple organ systems and histological classes, including brain tumours, mesenchymal neoplasms, renal tumours and germ cell tumours. TITAN-best diagnoses covered a broad performance range, from near-ceiling retrieval for paraganglioma (\(F1=0.99\) in both settings), clear cell renal cell carcinoma (\(F1=0.96\)) and seminoma (\(F1=0.91\) for Top-1; \(F1=0.90\) for Top-3), to more difficult entities such as astrocytoma (\(F1=0.27\) in both settings) and dedifferentiated liposarcoma (\(F1=0.44\) for Top-1; \(F1=0.39\) for Top-3). These findings indicate that TITAN's advantages were not restricted to a single organ system or to only high-performing diagnoses.

Further, we quantified diagnosis-level rank-1 wins, defined as the number of diagnostic categories for which each model achieved the highest F1-score (Fig.~\ref{fig:diagnosis_level_performance}\textbf{c,d}). Under Top-1 retrieval, TITAN ranked first in 20 diagnoses, followed closely by ABMIL in 15, Yottixel+CONCH (50\%) in 5, Madeleine in 4, Yottixel (5\%) in 3, PRISM and GigaPath in 2 each, and Yottixel+CONCH  (10\%) in 1; Yottixel+CONCH (100\%) and Yottixel+CONCH (20\%) did not rank first in any diagnosis. Thus, despite TITAN's superior aggregate mean performance, ABMIL alone captured 15 diagnosis-level wins, and competing models collectively led in a substantial fraction of diagnostic categories. Under Top-3 retrieval, TITAN led 19 diagnoses and ABMIL led 15, while the remaining wins were more broadly distributed: Yottixel+CONCH (100\%) and Yottixel+CONCH (20\%) each ranked first in 4 diagnoses, Madeleine in 4, PRISM in 3, Yottixel+CONCH (50\%) in 3, Yottixel+CONCH (10\%) in 2, GigaPath in 2, and Yottixel+CONCH (5\%) in 1. The gain of Yottixel+CONCH (100\%) and (20\%) under Top-3 (from 0 to 4 wins each) indicates that certain patch-sampling densities capture complementary morphological information that benefits from the expanded neighbourhood in majority-vote retrieval.

Collectively, these findings demonstrate that diagnosis-level retrieval performance is highly non-uniform and that model superiority depends strongly on diagnostic context. Aggregate organ-level metrics therefore obscure meaningful category-specific differences. Importantly, ABMIL's competitive showing should be interpreted with caution: as a supervised aggregator trained and evaluated within TCGA, it benefits from in-distribution optimization that is not available to the pre-trained foundation models operating in a zero-shot setting. Compared to MIL frameworks, TITAN is less amenable to fine-tuning and seems to be more commonly used as a fixed feature extractor, as effective adaptation typically requires substantial computational resources, careful optimization, and large annotated datasets—limitations that are exacerbated in clinical settings.
Within the strictly zero-shot configurations, Yottixel+CONCH at 100\% and 50\% sampling emerged as the strongest alternatives to TITAN across both Top-1 and Top-3 retrieval, matching or approaching TITAN's diagnosis-level win count in several contexts without any task-specific training. This suggests that for many diagnostic categories, dense patch-level retrieval with a strong encoder may be sufficient---and represents the more practically meaningful benchmark against which slide-level FMs should be assessed in deployment scenarios where labelled in-domain data are unavailable.

\subsection{Structured Misclassification Patterns in Retrieval}
We examined misclassification patterns for the lowest-support diagnoses (5--7 cases each; see Methods) across all ten retrieval configurations under both Top-1 and Top-3 retrieval (Extended Data Fig.~\ref{fig:misclassification_top1}, ~\ref{fig:misclassification_top3}). Under Top-1 retrieval, five diagnoses exhibited universal misclassification across all ten models; under Top-3 retrieval, this reduced to three, with combined hepatocellular--cholangiocarcinoma and yolk sac tumour no longer appearing as complete failures. Across both retrieval depths, misclassifications did not scatter randomly but were concentrated in a limited set of alternative classes. Errors were predominantly directed toward morphologically related histologic categories within the same organ: combined hepatocellular–cholangiocarcinoma cases were most frequently retrieved as hepatocellular carcinoma or cholangiocarcinoma; diffuse sclerosing variants of papillary thyroid carcinoma were primarily misclassified as canonical papillary carcinoma or columnar cell variants; and yolk sac tumour misclassification targets under Top-1 were confined to other germ cell subtypes. Under Top-3 retrieval, the three persistently failing diagnoses showed a marginally broader misclassification spectrum — for instance, Breast/Medullary carcinoma acquired metaplastic carcinoma and ductal and lobular carcinoma as additional error targets beyond invasive carcinoma subtypes — yet all errors remained strictly organ-confined. Across models, misclassification distributions varied in specific label frequencies, but the overall pattern of concentrated, organ-confined errors was consistent across all ten retrieval configurations, including the newly added ABMIL and Yottixel+CONCH (100\%), which did not introduce cross-organ misclassification for any of the diagnoses shown.

\subsection{Prevalence shapes but does not fully explain diagnosis-level retrieval performance}

\begin{figure}[htbp]
    \centering
    \includegraphics[width=0.8\linewidth]{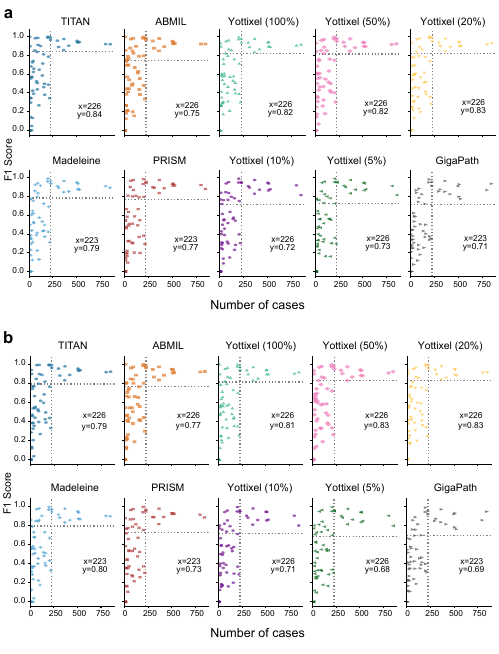}
    \caption{\textbf{Diagnosis prevalence is associated with, but does not fully determine, retrieval performance.} \textbf{a,b}, Scatter plots show diagnosis-level F1 score (y-axis) versus the number of cases for that diagnosis (x-axis) under Top-1 (\textbf{a}) and Top-3 (\textbf{b}) retrieval. Each point represents one diagnosis. Dotted reference lines indicate the Gaussian-mixture-derived case-count threshold ($x$) and F1 threshold ($y$) used to partition diagnoses into support-performance regimes for each model (values annotated within each subplot; see Methods). The case-count threshold was highly stable across models ($x = 223$--226 in all panels). Top-1 F1 thresholds were 0.84 (TITAN), 0.75 (ABMIL), 0.82 (Yottixel+CONCH (100\%)), 0.82 (Yottixel+CONCH (50\%)), 0.83 (Yottixel+CONCH (20\%)), 0.79 (Madeleine), 0.77 (PRISM), 0.72 (Yottixel+CONCH (10\%)), 0.73 (Yottixel+CONCH (5\%)) and 0.71 (GigaPath). Top-3 F1 thresholds were 0.79, 0.77, 0.81, 0.83, 0.83, 0.80, 0.73, 0.71, 0.68 and 0.69, respectively. [Note that Yottixel+CONCH has been shortened to Yottixel in the figure]}
    \label{fig:f1_score_rare_case}
\end{figure}

To assess whether class prevalence drives diagnosis-level retrieval performance, we examined the relationship between diagnosis support and F1-score using linear-scale scatter plots (Fig.~\ref{fig:f1_score_rare_case}). Across TITAN, ABMIL, Yottixel+CONCH variants, Madeleine, PRISM and GigaPath, diagnosis support was positively associated with F1-score, indicating that common diagnoses tended to be retrieved more accurately than rare diagnoses. At the same time, prevalence alone did not explain the full performance landscape. Several diagnoses with comparable support showed markedly different retrieval accuracy, indicating that intrinsic morphological distinctiveness and overlap between related entities also shaped retrieval difficulty.

Despite this overall trend, substantial dispersion was evident at all support levels, with multiple rare diagnoses achieving near-perfect performance and some relatively common diagnoses remaining challenging. Across models, performance generally stabilized for diagnoses exceeding approximately 200 cases, where F1-scores concentrated near 0.9--1.0 with limited variance. In contrast, diagnoses below this threshold exhibited considerable heterogeneity in retrieval performance. Notably, the data-driven thresholds used to segment diagnoses were highly similar across models and retrieval settings: the case-count split was nearly invariant (\(x = 223\)--226 for all models in both Top-1 and Top-3 retrieval), whereas the F1 threshold varied only modestly (\(y = 0.71\)--0.84 in Top-1 and \(y = 0.68\)--0.83 in Top-3). This consistency suggests that the broad prevalence--performance structure is stable across embedding spaces and top-\(n\) retrieval definitions. These thresholds highlight three recurring regimes: (i) low-support, low-performance rare failures; (ii) low-support, high-performance rare-but-easy diagnoses; and (iii) high-support, high-performance diagnoses for which retrieval is consistently strong. Together, these results indicate that prevalence contributes substantially to diagnosis-level retrieval performance but does not fully explain it, suggesting that morphological ambiguity, intra-class heterogeneity and label granularity also shape retrieval success.

\clearpage
\section{Discussion}\label{sec3}

Across 9{,}387 TCGA diagnostic slides spanning 17 organs and 60 organ-specific diagnoses, our benchmark reveals that the choice of retrieval architecture matters far less than the diagnostic context in which it is deployed. TITAN emerged as the strongest pre-trained slide foundation model overall, yet its advantage was modest: under Top-1 retrieval it was statistically indistinguishable from ABMIL, and under Top-3 retrieval multiple Yottixel configurations likewise reached parity. Importantly, ABMIL is a supervised aggregator trained and evaluated within TCGA, so its competitive performance reflects in-distribution optimization rather than zero-shot generalization; that TITAN matched this in-distribution baseline without any task-specific training underscores the quality of its pre-trained representations. Beneath these aggregate statistics, performance varied far more across organs and diagnoses than across models, with morphologically distinctive entities approaching ceiling retrieval regardless of pipeline and rare or overlapping subtypes remaining universally difficult---a pattern consistent with the pronounced organ-level variability reported in prior retrieval studies~\cite{alfasly_validation_2025,kalra_pan-cancer_2020,lahr_analysis_2025}. No single model dominated all 17 organs or all 60 diagnostic categories; instead, organ- and diagnosis-level leadership was distributed across TITAN, ABMIL, Yottixel variants, Madeleine, PRISM and GigaPath, with different pipelines excelling for different subsets of diagnoses. This dispersed competitive landscape echoes recent benchmarking findings that no single aggregation strategy or foundation model dominates across tasks~\cite{chen_towards_2024,neidlinger_benchmarking_2025,campanella_clinical_2025}, and argues against reliance on any one retrieval architecture. Instead, it supports ensemble or retrieval-fusion strategies that leverage complementary model strengths~\cite{neidlinger_benchmarking_2025}---an approach further motivated by the broader set of competitive cases observed under Top-3 retrieval, where multi-candidate frameworks preserve diagnostically relevant alternatives even when the top-ranked slide is incorrect.
 
The observed misclassification patterns suggest that the embedding spaces preserve meaningful organ- and lineage-level structure but remain insufficiently resolved for certain fine-grained subtype distinctions. Incorrect retrievals were typically concentrated among biologically coherent neighbouring classes rather than being randomly distributed, implying that current models capture broad morphological similarity even when they fail at exact subtype recovery. The tendency for errors to drift towards common or dominant subtypes within an organ further points to the joint influence of data density and morphological prevalence on retrieval behaviour. Together, these findings indicate that current FMs encode high-level histopathological organisation effectively, but remain limited in their ability to separate subtle variant morphology, a constraint that should guide both future model development and benchmark design. Misclassifications were highly structured, typically confined to biologically and morphologically adjacent classes, indicating that current models capture high-level histopathological organization but lack sufficient resolution for fine-grained subtype discrimination. Notably, these failure modes align with areas of known inter-observer variability among expert pathologists, suggesting an intrinsic ceiling for morphology-only retrieval and motivating integration of multimodal data. 
 
Importantly, the diagnostic boundaries at which retrieval performance is lowest correspond closely to those at which human expert pathologists also exhibit poor interobserver agreement, suggesting that the observed failures reflect intrinsic morphological ambiguity in H\&E-stained tissue rather than model-specific representational deficiencies. In thymus---the lowest-performing organ in our benchmark---an international panel of 13 expert pathologists classifying 305 thymic epithelial tumours under the same WHO subtypes present in our dataset achieved only moderate agreement ($\kappa = 0.49$) at the clinically actionable boundary between B3 and the remaining thymoma subtypes, with the principal confusions---B1 versus B2 versus B3, and B3 versus thymic carcinoma---mapping directly onto the structured misclassification patterns we observed~\cite{wolf_interobserver_2020}. In thyroid, 14 subspecialist pathologists achieved only fair agreement ($\kappa = 0.34$) on the tall cell variant of papillary thyroid carcinoma, with four distinct diagnostic definitions in concurrent use and 68\% of classical PTC cases reclassified by at least one panelist~\cite{hernandez-prera_pathologic_2018}---a degree of human diagnostic uncertainty that parallels our finding of strictly intra-family misclassification among papillary carcinoma variants. In brain, a population-based glioma series documented 23\% overall discordance between initial and review diagnoses, rising to 62\% for astrocytoma and 42--52\% for oligodendroglial tumours~\cite{aldape_discrepancies_2000}, diagnostic distinctions that likewise proved resistant to accurate retrieval across all ten pipelines. This convergence between computational retrieval failures and human diagnostic disagreement implies that the performance floor we observe for these organs may approximate a practical ceiling for purely morphology-based approaches, and that further gains will likely require integration of molecular and immunohistochemical ancillary data into multimodal retrieval frameworks.
 
These findings resonate with recent critiques arguing that current FMs in pathology remain misaligned with practical clinical needs, citing low retrieval accuracy, fragile cross-institutional generalization and sensitivity to routine imaging variation as persistent shortcomings \cite{tizhoosh_beyond_2025}. Our organ- and diagnosis-resolved analysis provides empirical support for several of these concerns: retrieval accuracy for rare and morphologically overlapping entities remains well below clinical thresholds across all ten evaluated pipelines, and the absence of a universally dominant model confirms that current architectures have not yet converged on robust, generalizable representations---a conclusion also supported by the diversity of model architectures and evaluation protocols documented in recent surveys~\cite{xiong_survey_2025}. At the same time, the structured nature of misclassification errors---concentrated within organ and lineage boundaries rather than scattered randomly---suggests that these models do capture meaningful histopathological organization, even where exact subtype discrimination fails. Our results thus refine the critique by localizing the failure modes: the primary bottleneck is not a wholesale absence of morphological encoding, but rather insufficient resolution at the level of fine-grained diagnostic subtypes and rare entities.
 
The strong showing of Yottixel further suggests that retrieval quality is presently driven more by feature quality than by sophisticated aggregation. Yottixel relies on a simple median-of-minimums distance over patch embeddings~\cite{kalra_yottixel_2020,kalra_pan-cancer_2020}, yet it remained highly competitive, particularly under Top-3 retrieval. Together with the performance of ABMIL~\cite{ilse_attention-based_nodate} built on CONCH v1.5 patch embeddings, this points to the feature extractor as the main determinant of retrieval quality in the current benchmark, with slide-level aggregation contributing comparatively little additional gain---a finding consistent with Alfasly et al., who observed limited benefit from GigaPath's slide-level aggregation over patch-level baselines~\cite{alfasly_validation_2025}, and with Chen et al., who found that no single aggregation strategy dominates across clinical tasks~\cite{chen_towards_2024}. This interpretation is also consistent with the broader pattern that poorer feature extractors substantially depress performance across otherwise capable pipelines.
 
Diagnosis-level analysis adds an important practical implication. Different models displayed different diagnosis-specific strengths, and some achieved near-perfect or perfect F1 scores for selected entities while performing less well on others. At the same time, the most difficult categories were typically rare, heterogeneous or morphologically overlapping diagnoses, with errors remaining largely confined to organ- or lineage-adjacent classes. These results argue against the expectation that one retrieval model will be uniformly optimal in every clinical context~\cite{alfasly_validation_2025,tizhoosh_beyond_2025}. A more realistic near-term direction may be organ-specific or diagnosis-aware deployment, in which retrieval systems are specialized, selected or ensembled according to tissue type and diagnostic question~\cite{lahr_analysis_2025}.

There are some limitations of this study. The comparison between TITAN and ABMIL should be interpreted in the context of training regime. ABMIL is supervised and evaluated within TCGA, and therefore benefits from in-domain optimization on the same dataset family used for testing. Its strong performance is consequently informative, but not unexpected. Against this baseline, TITAN's competitiveness is notable because it reflects the behaviour of a pre-trained slide foundation model used in a zero-shot retrieval setting. That gap in supervision makes TITAN's performance less a sign that slide FMs have failed to surpass supervised aggregation and more a sign that such models can transfer surprisingly well to an unseen retrieval task without retraining.

\section{Conclusion}\label{sec4}
This benchmark of ten WSI retrieval pipelines across 9{,}387 TCGA slides, 17 organs and 60 diagnostic categories establishes that current FMs for histopathology retrieval are not yet converging on a single dominant architecture. TITAN was the strongest pre-trained slide-level model overall, yet its margin over both a supervised in-distribution aggregator (ABMIL) and dense patch-level retrieval (Yottixel at high sampling rates) was narrow and often non-significant. Performance heterogeneity across organs and diagnoses consistently exceeded differences across models, indicating that the morphological complexity of the diagnostic landscape---not the retrieval pipeline---is the primary determinant of accuracy. Misclassification errors were structured rather than random, concentrating within organ and lineage boundaries, which suggests that current embeddings capture broad histopathological organization but lack the resolution needed for fine-grained subtype discrimination. Runtime analysis showed that patch encoding dominated end-to-end cost, with slide-level aggregation contributing minimally to total inference time. These findings collectively argue for organ-resolved benchmarking as a standard evaluation practice, stronger patch-level representations as the most impactful avenue for improvement, and hybrid retrieval strategies that combine complementary slide- and patch-level signals to address the diverse diagnostic contexts encountered in clinical deployment.

\section{Methods}\label{sec5}
\subsection{Dataset}

We constructed our cohort from TCGA diagnostic whole-slide histopathology images \cite{weinstein2013cancer}. The dataset comprises 11,765 hematoxylin and eosin (H\&E) whole-slide images, corresponding to 9,640 unique patients and spanning 25 distinct organs. The original TCGA primary diagnoses were not designed for FM benchmarking and present several challenges: categories are not always mutually exclusive, some terminology is outdated relative to current World Health Organization classifications, and certain distinctions—such as keratinizing versus non-keratinizing squamous cell carcinoma subtypes—reflect criteria with documented poor inter-observer reproducibility \cite{uegami2026}. To address these limitations, we adopted the standardized WSI classification scheme proposed by Uegami et al.\ \cite{uegami2026}, which systematically re-evaluates TCGA diagnoses across organ systems by merging ambiguous or morphologically indistinguishable subcategories, updating outdated nomenclature to align with current World Health Organization standards, and excluding categories unsuitable for image-based machine learning classification. This refined labeling scheme reduces semantic ambiguity and improves label mutual exclusivity, yielding a final cohort of 9,387 diagnostic slides spanning 17 organs and 60 organ-specific diagnostic categories. Because histologic entities with identical names may represent biologically and clinically distinct diseases depending on anatomical context, we defined a diagnostic category as the combination of Organ × diagnosis (Extended Data Table ~\ref{tab:shared_labels}: All retrieval experiments were conducted using this organ-specific diagnostic definition.

\subsubsection{Exclusion criteria}
To ensure statistically valid patient-level retrieval evaluation, we applied the following exclusion criteria prior to model evaluation:
\begin{itemize}
\item Minimum patient count per diagnosis: 
Diagnostic categories represented by fewer than four unique patients were excluded. Because performance was evaluated using top-3 patient-level retrieval, each query case requires at least three distinct patients with the same diagnosis available in the retrieval database. Therefore, a minimum of four patients per diagnosis is necessary: one serving as the query case and at least three remaining cases to allow a correct top-3 match.

\item Minimum diagnostic diversity per organ: 
Retrieval evaluation was performed vertically within each organ (i.e., cases were compared only against other cases from the same organ). To compute meaningful F1 scores per-diagnosis, each organ was required to contain at least two distinct diagnostic categories. Organs with only a single diagnosis type were excluded, as retrieval in such cases would be trivial and would not allow the computation of class-specific precision, recall, or F1 metrics.
\end{itemize}
These criteria ensured that (i) top-3 retrieval was mathematically feasible for every diagnosis, and (ii) evaluation metrics reflected true intra-organ diagnostic discrimination rather than organ-level separation. The exclusion of organs and diagnoses that did not meet the minimum patient count and diagnosis count yielded 17 organs and 60 diagnoses (Extended Data Table ~\ref{tab:nested_diagnosis}, also see the white paper with TCGA subtype information \cite{uegami2026}).

\subsection{Embedding extraction and retrieval paradigm}
WSIs were processed using the Trident platform \cite{zhang2025standardizing}, for patch extraction, patch-level embedding processing, and slide-level embedding generation. WSIs were tiled into fixed-size patches following tissue masking and background filtering. Patch-level embeddings were generated using pretrained transformer-based patch encoders. These encoders were used in inference mode without additional fine-tuning on TCGA. Slide-level embeddings were computed by aggregating patch-level representations into a single vector per slide using slide-level FMs. Both patch-level and slide-level embeddings were retained for downstream retrieval evaluation.
\subsubsection{Models used for evaluation}
 To ensure a fair evaluation and avoid data leakage, benchmarking on TCGA was restricted to FMs whose pretraining corpora did not include TCGA data. Under this constraint, we evaluated five slide-level models: four publicly available pre-trained histopathology FMs---Titan \cite{ding2025multimodal}, Madeleine \cite{leonardis_multistain_2025}, PRISM \cite{shaikovski_prism_2024}, and Gigapath \cite{xu_whole-slide_2024}--and one supervised aggregator, ABMIL (described below). We also evaluated a high-performing patch-level encoder, Conch v1.5 \cite{lu_visual-language_2024}, identified as one of the top patch encoders in a recent large-scale benchmark \cite{neidlinger_benchmarking_2025}. Conch v1.5 also serves as the feature extractor backbone within Titan, ABMIL and Yottixel \cite{kalra_yottixel_2020}, enabling direct comparison between unsupervised pre-trained aggregation (TITAN), supervised aggregation (ABMIL), and aggregation-free patch-level retrieval---all built upon the same patch representations. The four pre-trained FMs were pretrained on large-scale, non-TCGA whole-slide image collections using self-supervised and/or multi-modal contrastive learning objectives and provide fixed patch-level encoders applied in inference mode without task-specific fine-tuning.
\subsubsection{Retrieval paradigm}
For slide-level FMs, each whole-slide image (WSI) was represented by a single fixed-dimensional embedding extracted in inference mode. Given a query WSI from the TCGA cohort, we computed pairwise Euclidean distances between its embedding and the embeddings of all other WSIs in the dataset. To prevent information leakage at the patient level, we adopted a leave-one-patient-out protocol: when a WSI from a given patient was used as the query, all WSIs from the same patient were excluded from the candidate retrieval pool. For each query, the top-n nearest neighbors (based on smallest Euclidean distance) were retrieved. Retrieval performance was evaluated using top-n majority voting. Specifically, for majority-of-top-n, the predicted diagnosis of the query slide was determined by the most frequent diagnosis label among the top-n retrieved WSIs. Performance was aggregated across slides and reported at the patient level to reflect clinically meaningful retrieval behavior.

We also trained an attention-based multiple instance learning (ABMIL) model \cite{ilse_attention-based_nodate} on top of frozen, pre-trained patch embeddings to measure how much performance improves when adding slide-level aggregation beyond the patch encoder alone. In this setup, each WSI is represented as a collection (or “bag”) of patch-level features extracted by a pre-trained model. The ABMIL model then learns to assign importance (attention weights) to each patch and combines them into a single slide-level representation, which is used to predict the diagnosis. To evaluate the performance of the ABMIL, we employed a patient-level 3-fold cross-validation protocol with grouped stratification by diagnosis. In each run, two folds were used for training and one fold was held out for testing. If a whole-slide image (WSI) from a given patient was assigned to the held-out fold, all other samples from that patient were excluded from the corresponding training set. Stratification was performed on the diagnosis labels to maintain a balanced class distribution across folds. Across folds, a given patient may contribute to training in one run and to testing in another, as is standard in cross-validation. To calculate the F1 score of the ABMIL, we pooled the retrieval results of the three test folds, yielding one F1 score for all three folds for organ-level and diagnosis-level retrieval, respectively.

As mentioned before, we needed a search engine that could easily integrate different FMs and process WSIs. Unsupervised patch selection is also preferred. Storage efficiency would be a highly desirable attribute of such a search engine. Based on these criteria, we selected Yottixel as our evaluation platform~\cite{tizhoosh2024image,kalra_yottixel_2020,lahr2024analysis}. Hence, we employed the Yottixel framework to generate a diversified mosaic of representative patches per WSI~\cite{kalra_yottixel_2020}. Each WSI is first segmented into chromatically distinct regions via $k$-means clustering, from which a fixed percentage of spatially diverse patches is sampled through a second round of location-based clustering, yielding a mosaic approximately 20-fold smaller than the full tissue area. Patch features were extracted using CONCH v1.5 and subsequently binarised into compact binary codes (barcodes) using the MinMax algorithm~\cite{tizhoosh_minmax_2016}; the collection of barcodes across all mosaic patches constitutes the Bunch of Barcodes (BoB) index for each WSI. Given two WSIs, patch-level similarity was computed using the median-of-minimum distances scheme~\cite{alfasly_validation_2025}. Specifically, for each barcode in the query WSI's BoB, the minimum Hamming distance to any barcode in the candidate WSI's BoB was identified; the median of these per-patch minimum distances served as the slide-to-slide dissimilarity score. This strategy captures local morphological correspondence while maintaining robustness to outliers and patch heterogeneity.

As in both slide-level and patch-level setting, a leave-one-patient-out protocol was enforced to eliminate intra-patient retrieval bias. For each query WSI, the top-n nearest slides were retrieved according to the computed patch-based distance, and diagnosis prediction was obtained via majority voting among the top-n neighbors. 

\subsection{Computation resources}

For embedding extraction and retrieval processing, experiments were conducted on a dual-socket Intel Xeon Gold 6542Y system (96 logical CPUs) equipped with NVIDIA H100 80GB HBM3 GPUs.

\subsection{Misclassification analysis}
To characterize structured error patterns, we selected the lowest-support diagnoses in the filtered cohort, defined as diagnostic categories represented by only 5--7 cases. For each such diagnosis, stacked bar plots were generated showing the distribution of incorrect retrieval labels produced by each of the ten retrieval configurations under both Top-1 and Top-3 retrieval. This analysis was restricted to misclassified cases (i.e., cases for which the retrieved diagnosis did not match the true label).

\subsection{Statistical comparisons and multiple-testing correction}
All pairwise comparisons of organ-level F1-macro scores between TITAN and each of the nine alternative retrieval configurations were performed using two-sided paired $t$-tests ($n = 17$ organs per comparison). Because nine simultaneous hypotheses were tested against a single reference model, we controlled the family-wise error rate using the Holm--Bonferroni sequential correction procedure \cite{abdi2010holm}. Specifically, the nine raw $P$-values were ranked in ascending order and compared against progressively relaxed significance thresholds $\alpha / (m - k + 1)$, where $\alpha = 0.05$, $m = 9$ is the number of comparisons and $k$ is the rank index. The procedure was applied independently for Top-1 and Top-3 retrieval settings. A comparison was deemed significant only if its ordered $P$-value fell below the corresponding Holm--Bonferroni threshold, ensuring that the probability of one or more false rejections across the family of tests remained below $\alpha$. 

\subsection{Gaussian mixture modeling to identify $F1$ score clusters}
This section describes the method used to separate the the F1 score clusters.
We used Gaussian mixture modelling (GMM) to derive objective, data-driven cut-offs that segment F1 scatter plots into interpretable performance regimes based on the case count. For each axis (case count and F1-score), a two-component univariate GMM was fitted to the observed distribution, and the threshold was defined as the intersection point at which the two weighted Gaussian densities are equal (i.e., equal posterior likelihood of belonging to either subpopulation). These thresholds were applied as orthogonal vertical and horizontal dividers to partition the scatter space into four quadrants (low vs high support; low vs high F1), enabling consistent identification of rare-but-high-performing and rare-and-poor-performing diagnoses without ad hoc parameter selection. To ensure robustness and reproducibility, model fitting used multiple initializations with a fixed random seed.
\backmatter

\bmhead{Supplementary information}

See Appendix A. 

\bmhead{Acknowledgments}
The authors gratefully acknowledge the \emph{F. Craig and Patricia Jilk Fund for Data Science, Predictive Modeling \& AI for Breast Cancer} for supporting this study.
The authors also acknowledge the \emph{Mayo Clinic Comprehensive Cancer Center}, Rochester, MN, USA, for its ongoing support.

\bmhead{Disclosures} Dr. Judy Boughey receives research funding paid to her institution from Eli Lilly, SimBioSys and Quantum Leap Healthcare. She is on the DSMB for Cairns Surgical.
Dr. Boughey is supported by the W.H. Odell Professorship in Individualised Medicine

\begin{appendices}

\clearpage
\section{Extended Data}\label{secA2}

\begin{figure}[htbp]
    \centering
    \includegraphics[width=\linewidth]{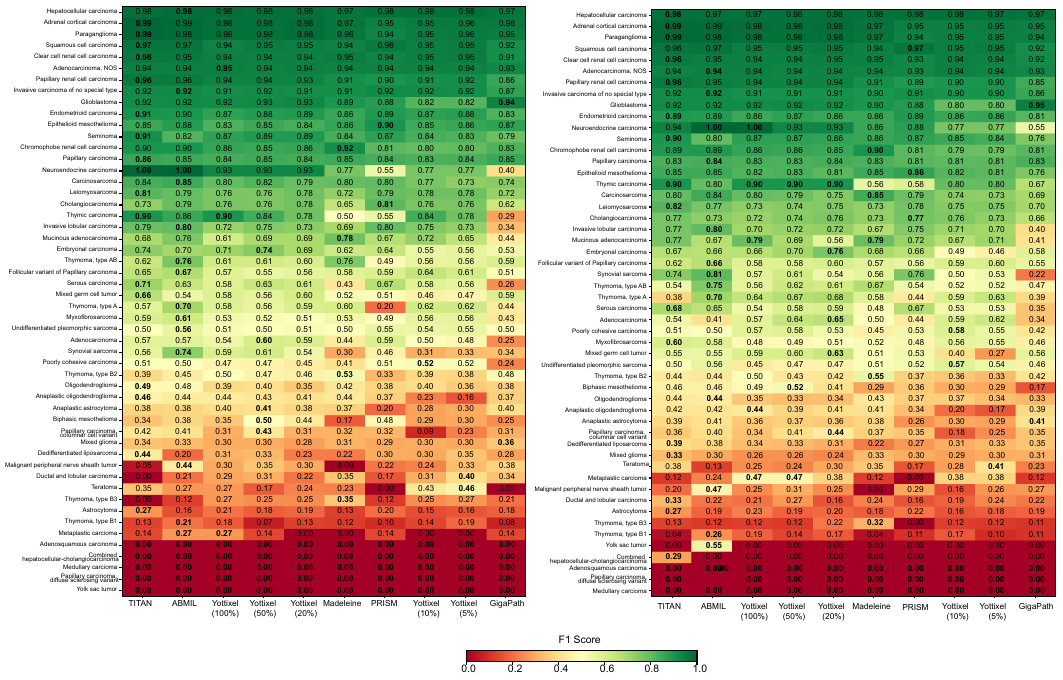}
    \caption{Complete diagnosis-level retrieval performance across all FMs and retrieval settings. Heatmaps show F1-macro scores for all 60 diagnoses under Top-1 (left) and Top-3 (right) retrieval. Rows correspond to diagnoses (ordered by mean F1 across models), and columns denote the ten evaluated retrieval configurations. Color intensity reflects absolute F1 score (0--1 scale). High-performing diagnoses cluster at the top of each panel, whereas rare and morphologically complex entities concentrate in the low-performance region at the bottom. }
    \label{fig:f1_diagnosis_breakdown}
\end{figure}

\begin{figure}[htbp]
    \centering
    \includegraphics[width=\linewidth]{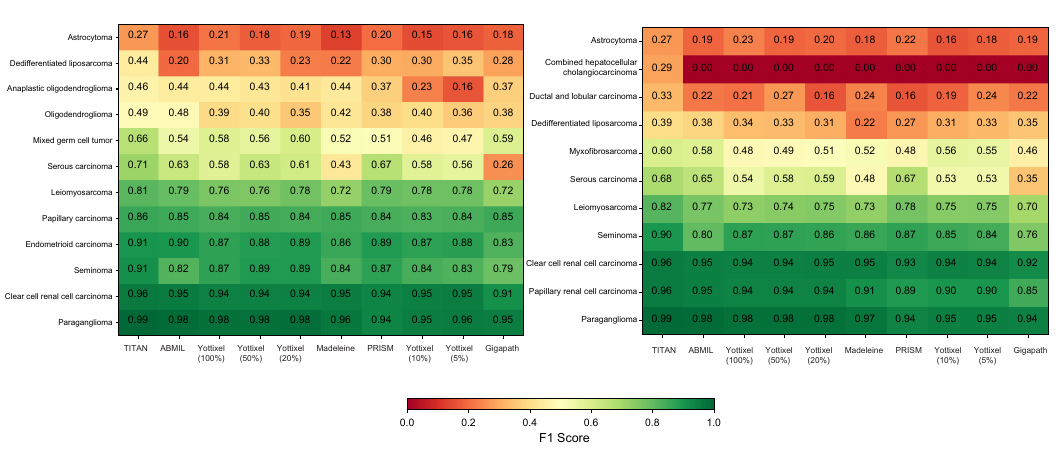}
    \caption{{Diagnosis-level F1-macro scores for diagnoses in which TITAN achieves the highest retrieval performance.} Heatmaps show per-model F1-macro scores for diagnoses in which TITAN is the unique top-performing model after rounding to two decimal places. Left, Top-1 retrieval (\(n=12\) diagnoses). Right, Top-3 retrieval (\(n=11\) diagnoses). Rows are ordered by TITAN F1-macro in ascending order, and columns show the ten evaluated models in overall rank order. Cell colours indicate absolute F1-macro values from 0 to 1, and bold text denotes the highest value in each row. TITAN-best diagnoses span highly retrievable entities, including paraganglioma and clear cell renal cell carcinoma, intermediate-performing entities such as serous carcinoma and leiomyosarcoma, and challenging diagnoses such as astrocytoma and dedifferentiated liposarcoma. The Top-3 panel additionally includes combined hepatocellular-cholangiocarcinoma, for which TITAN retains non-zero retrieval performance whereas all other models remain at zero.} 
    \label{fig:titan_best_diagnoses}
\end{figure}

\begin{figure}[htbp]
    \centering
    \includegraphics[width=\linewidth]{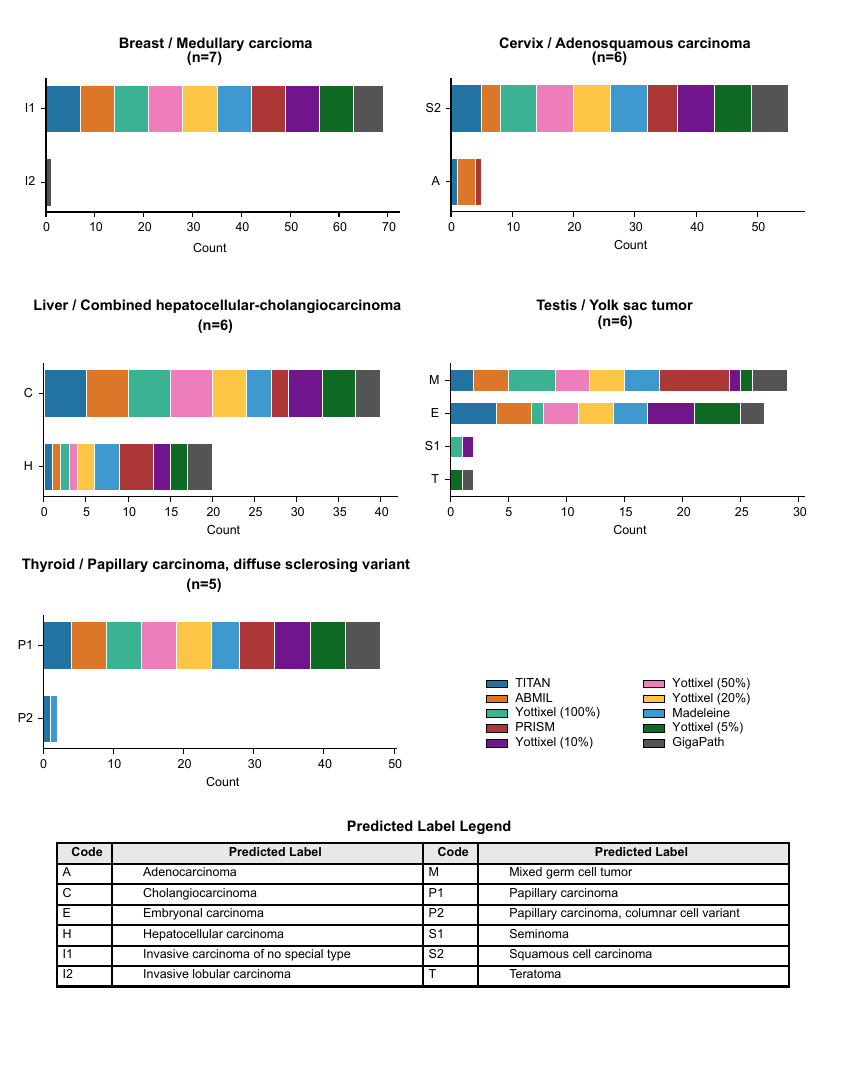}
    \caption{Misclassification targets of the diagnoses under Top-1 retrieval. Stacked bar plots show the distribution of incorrect predicted labels for Breast/Medullary carcinoma (n=7), Cervix/Adenosquamous carcinoma (n=6), Liver/Combined hepatocellular-cholangiocarcinoma (n=6), Testis/Yolk sac tumor (n=6), and Thyroid/Papillary carcinoma diffuse sclerosing variant (n=5). Each panel title indicates the true organ--diagnosis pair and case count. Bars show the cumulative count of incorrect labels retrieved for all ten retrieval configurations: TITAN, ABMIL, Yottixel (100\%, 50\%, 20\%, 10\%, 5\%), Madeleine, PRISM and GigaPath. Only misclassified cases are included. Bar colors denote individual models (see model legend); predicted label codes are decoded in the Predicted Label Legend table below.}
    \label{fig:misclassification_top1}
\end{figure}

\begin{figure}[htbp]
    \centering
    \includegraphics[width=\linewidth]{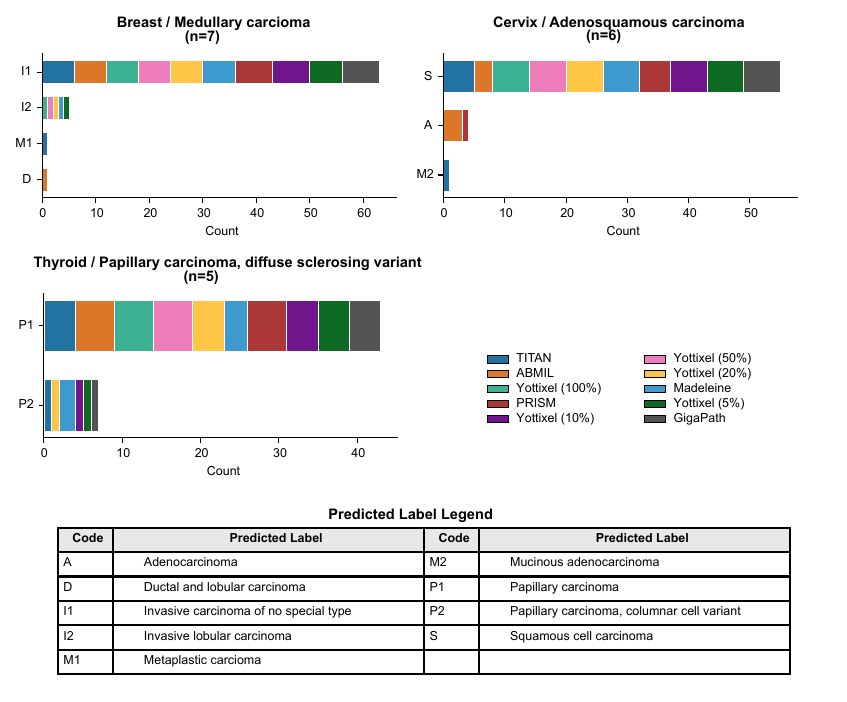}
    \caption{Misclassification targets of the diagnoses under Top-3 retrieval. Stacked bar plots show the distribution of incorrect predicted labels for Breast/Medullary carcinoma (n=7), Cervix/Adenosquamous carcinoma (n=6), and Thyroid/Papillary carcinoma diffuse sclerosing variant (n=5) . Each panel title indicates the true organ--diagnosis pair and case count. Bars show the cumulative count of incorrect labels retrieved for all ten retrieval configurations: TITAN, ABMIL, Yottixel (100\%, 50\%, 20\%, 10\%, 5\%), Madeleine, PRISM and GigaPath. Only misclassified cases are included. Bar colors denote individual models (see model legend); predicted label codes are decoded in the Predicted Label Legend table below.}
    \label{fig:misclassification_top3}
\end{figure}

\begin{table}[ht]
\centering
\caption{Diagnostic labels appearing in multiple organs. Diagnoses were defined as organ-specific (Organ $\times$ Diagnosis).}
\label{tab:shared_labels}
\begin{tabular}{lll}
\toprule
\textbf{Diagnosis Label} & \textbf{\# Organs} & \textbf{Organs} \\
\midrule
Adenocarcinoma & 7 & Colon; Rectum; Stomach; Esophagus; Lung; Pancreas; Prostate \\
Squamous Cell Carcinoma & 4 & Lung; Esophagus; Cervix; Head and Neck \\
Clear Cell Carcinoma & 3 & Kidney; Ovary; Uterus \\
Mucinous Adenocarcinoma & 3 & Colon; Rectum; Ovary \\
Papillary Carcinoma & 2 & Thyroid; Kidney \\
Serous Carcinoma & 2 & Ovary; Uterus \\
Neuroendocrine Tumor / Carcinoma & 3 & Lung; Pancreas; Gastrointestinal tract \\
Sarcoma & 3 & Soft tissue; Uterus; Bone \\
Lymphoma & 3 & Lymph node; Spleen; Gastrointestinal tract \\
\bottomrule
\end{tabular}
\end{table}

\begin{table}[ht]
\centering
\caption{Diagnosis breakdown by organ.}
\label{tab:nested_diagnosis}
\begin{tabular}{llccc}
\toprule
\textbf{Organ} & \textbf{Total WSIs} & \textbf{Diagnosis} & \textbf{WSIs} & \textbf{\% of Organ} \\
\midrule

\multirow{2}{*}{Adrenal} 
& \multirow{2}{*}{420} 
& Adrenal cortical carcinoma & 224 & 53.3\% \\
& & Paraganglioma & 196 & 46.7\% \\
\midrule

\multirow{6}{*}{Brain} 
& \multirow{6}{*}{1689} 
& Anaplastic astrocytoma & 164 & 9.7\% \\
& & Anaplastic oligodendroglioma & 155 & 9.2\% \\
& & Astrocytoma & 103 & 6.1\% \\
& & Glioblastoma & 850 & 50.3\% \\
& & Mixed glioma & 215 & 12.7\% \\
& & Oligodendroglioma & 202 & 12.0\% \\
\midrule

\multirow{6}{*}{Breast} 
& \multirow{6}{*}{1072} 
& Ductal and lobular carcinoma & 27 & 2.5\% \\
& & Invasive carcinoma of no special type & 802 & 74.8\% \\
& & Invasive lobular carcinoma & 203 & 18.9\% \\
& & Medullary carcinoma & 7 & 0.7\% \\
& & Metaplastic carcinoma & 13 & 1.2\% \\
& & Mucinous adenocarcinoma & 20 & 1.9\% \\
\midrule

\multirow{4}{*}{Cervix} 
& \multirow{4}{*}{277} 
& Adenocarcinoma & 28 & 10.1\% \\
& & Adenosquamous carcinoma & 6 & 2.2\% \\
& & Mucinous adenocarcinoma & 17 & 6.1\% \\
& & Squamous cell carcinoma & 226 & 81.6\% \\
\midrule

\multirow{2}{*}{Colorectum} 
& \multirow{2}{*}{590} 
& Adenocarcinoma, NOS & 512 & 86.8\% \\
& & Mucinous adenocarcinoma & 78 & 13.2\% \\
\midrule

\multirow{2}{*}{Esophagus} 
& \multirow{2}{*}{158} 
& Adenocarcinoma & 66 & 41.8\% \\
& & Squamous cell carcinoma & 92 & 58.2\% \\
\midrule

\multirow{3}{*}{Kidney} 
& \multirow{3}{*}{921} 
& Chromophobe renal cell carcinoma & 121 & 13.1\% \\
& & Clear cell renal cell carcinoma & 503 & 54.6\% \\
& & Papillary renal cell carcinoma & 297 & 32.2\% \\
\midrule

\multirow{3}{*}{Liver} 
& \multirow{3}{*}{408} 
& Cholangiocarcinoma & 37 & 9.1\% \\
& & Combined hepatocellular-cholangiocarcinoma & 6 & 1.5\% \\
& & Hepatocellular carcinoma & 365 & 89.5\% \\
\midrule

\multirow{2}{*}{Lung} 
& \multirow{2}{*}{1036} 
& Adenocarcinoma & 527 & 50.9\% \\
& & Squamous cell carcinoma & 509 & 49.1\% \\
\midrule

\multirow{2}{*}{Mesothelioma} 
& \multirow{2}{*}{80} 
& Biphasic mesothelioma & 19 & 23.8\% \\
& & Epithelioid mesothelioma & 61 & 76.2\% \\
\midrule

\multirow{2}{*}{Pancreas} 
& \multirow{2}{*}{207} 
& Adenocarcinoma & 199 & 96.1\% \\
& & Neuroendocrine carcinoma & 8 & 3.9\% \\
\midrule

\multirow{6}{*}{SoftTissue} 
& \multirow{6}{*}{587} 
& Dedifferentiated liposarcoma & 87 & 14.8\% \\
& & Leiomyosarcoma & 153 & 26.1\% \\
& & Malignant peripheral nerve sheath tumor & 27 & 4.6\% \\
& & Myxofibrosarcoma & 139 & 23.7\% \\
& & Synovial sarcoma & 20 & 3.4\% \\
& & Undifferentiated pleomorphic sarcoma & 161 & 27.4\% \\
\midrule

\multirow{2}{*}{Stomach} 
& \multirow{2}{*}{396} 
& Adenocarcinoma & 311 & 78.5\% \\
& & Poorly cohesive carcinoma & 85 & 21.5\% \\
\midrule

\multirow{5}{*}{Testis} 
& \multirow{5}{*}{209} 
& Embryonal carcinoma & 42 & 20.1\% \\
& & Mixed germ cell tumor & 50 & 23.9\% \\
& & Seminoma & 94 & 45.0\% \\
& & Teratoma & 17 & 8.1\% \\
& & Yolk sac tumor & 6 & 2.9\% \\
\midrule

\multirow{6}{*}{Thymus} 
& \multirow{6}{*}{179} 
& Thymic carcinoma & 11 & 6.1\% \\
& & Thymoma, type A & 27 & 15.1\% \\
& & Thymoma, type AB & 45 & 25.1\% \\
& & Thymoma, type B1 & 35 & 19.6\% \\
& & Thymoma, type B2 & 48 & 26.8\% \\
& & Thymoma, type B3 & 13 & 7.3\% \\
\midrule

\multirow{4}{*}{Thyroid} 
& \multirow{4}{*}{513} 
& Follicular variant of papillary carcinoma & 107 & 20.9\% \\
& & Papillary carcinoma & 363 & 70.8\% \\
& & Papillary carcinoma, columnar cell variant & 38 & 7.4\% \\
& & Papillary carcinoma, diffuse sclerosing variant & 5 & 1.0\% \\
\midrule

\multirow{3}{*}{Uterus} 
& \multirow{3}{*}{645} 
& Carcinosarcoma & 85 & 13.2\% \\
& & Endometrioid carcinoma & 416 & 64.5\% \\
& & Serous carcinoma & 144 & 22.3\% \\
\bottomrule
\end{tabular}
\end{table}



\end{appendices}

\clearpage
\bibliography{sn-bibliography}

\end{document}